\documentclass[conference]{IEEEtran}
\IEEEoverridecommandlockouts
\usepackage{hyperref}
\usepackage{amsmath,amssymb,amsfonts}
\usepackage{algorithmic}
\usepackage{algorithm}
\usepackage{graphicx}
\usepackage{textcomp}
\usepackage{xcolor}
\usepackage{subcaption}

\def\BibTeX{{\rm B\kern-.05em{\sc i\kern-.025em b}\kern-.08em
    T\kern-.1667em\lower.7ex\hbox{E}\kern-.125emX}}

\begin{document}

\bibliographystyle{unsrt}

\title{Avoidance Navigation Based on Offline Pre-Training Reinforcement Learning \\
\thanks{This research project is partially supported by the Ministry of Education, Singapore, under its Research Centre of Excellence award to the Institute for Functional Intelligent Materials (I-FIM, project No. EDUNC-33-18-279-V12). The simulation no collision navigation presentation will show in this link. \url{https://www.youtube.com/watch?v=fnSuDBk_iOY}}
}

\author{\IEEEauthorblockN{1\textsuperscript{st} Yang Wenkai}
\IEEEauthorblockA{\textit{National Unveristy of Singapore} \\
\textit{ECE Department of College of Design and Engineer}\\
Singapore \\
e0983030@u.nus.edu}
\and
\IEEEauthorblockN{2\textsuperscript{nd} Ji Ruihang}
\IEEEauthorblockA{\textit{National Unveristy of Singapore} \\
\textit{ECE Department of College of Design and Engineer}\\
Singapore \\
}
\and
\IEEEauthorblockN{3\textsuperscript{rd} Zhang Yuxiang}
\IEEEauthorblockA{\textit{National Unveristy of Singapore} \\
\textit{ECE Department of College of Design and Engineer}\\
Singapore \\
}
\and
\IEEEauthorblockN{4\textsuperscript{th} Lei Hao}
\IEEEauthorblockA{\textit{National Unveristy of Singapore} \\
\textit{ECE Department of College of Design and Engineer}\\
Singapore \\
}
\and
\IEEEauthorblockN{5\textsuperscript{th} Zhao Zijie}
\IEEEauthorblockA{\textit{National Unveristy of Singapore} \\
\textit{ECE Department of College of Design and Engineer}\\
Singapore \\
}
}

\maketitle

\begin{abstract}
This paper presents a Pre-Training Deep Reinforcement Learning(DRL) for avoidance navigation without map for mobile robots which map raw sensor data to control variable and navigate in an unknown environment. The efficient offline training strategy is proposed to speed up the inefficient random explorations in early stage and we also collect a universal dataset including expert experience for offline training, which is of some significance for other navigation training work. The pre-training and prioritized expert experience are proposed to reduce 80\% training time and has been verified to improve the 2 times reward of DRL. The advanced simulation gazebo with real physical modelling and dynamic equations reduce the gap between sim-to-real. We train our model a corridor environment, and evaluate the model in different environment getting the same effect. Compared to traditional method navigation, we can confirm the trained model can be directly applied into different scenarios and have the ability to no collision navigate. It was demonstrated that our DRL model have universal general capacity in different environment.
\end{abstract}

\begin{IEEEkeywords}
offline DRL, no collision, no map navigation, pre-training, prioritized expert experience
\end{IEEEkeywords}

\section{Introduction}
Traditional navigation methods, such as probabilistic reasoning\cite{ziebart2008navigate} and graph-based algorithms \cite{fu2004practical} \cite{herman2000graph}, often fail in complex, dynamic, and uncertain environments that require pre-built maps. While Deep Reinforcement Learning (DRL)\cite{li2017deep} \cite{franccois2018introduction} could solve these problems via deep neural network and is increasingly being adopted for mobile robot navigation \cite{desouza2002vision}, impacting various aspects of daily life, including self-driving cars, delivery drones, and personal robotic assistants.

DRL, combining deep learning's ability to handle high-dimensional sensory data and reinforcement learning's optimal decision-making\cite{kennerley2006optimal} capabilities, offers a robust framework for learning navigation policies directly from raw sensory input.

However, DRL's implementation in mobile robot navigation has challenges without map\cite{gaussier1997visual}. Unlike supervised learning\cite{kahn2021badgr}, with its mature datasets and defined tasks, DRL involves an intricate and resource-intensive training process. DRL agents need to interact with their environments to collect state-action-reward data, which can be both time-consuming and computationally demanding. In the initial training phase, the agent's random exploration\cite{tijsma2016comparing} often lead to local optima, inhibiting efficient learning.

To address these issues, we propose a novel approach to mobile robot navigation using offline Reinforcement Learning\cite{levine2020offline} based on laser data. This method introduces a new concept - a universal dataset specifically designed for mobile robot navigation\cite{rudenko2020thor}. Unlike traditional online DRL methods, our approach does not rely on simulation, thereby significantly reducing the training time. In this paper, we will delve into the details of this dataset and our offline training methodology, illuminating their potential to enhance the efficiency and effectiveness of mobile robot navigation using DRL.

Firstly, we will undertake a detailed exploration of our methodology, providing insights into the data collection process and the specifics of our DRL framework. The traditional path planning techniques as well as the use of the Robot Operating System (ROS) are elucidate for data gathering. Subsequently, we will turn our attention to the core of our study - our DRL framework. And we will present an in-depth discussion of its architecture, algorithms, and implementation details. The collected offline dataset will be used to train a pre-trained model which is fine-tuned on our simulation. During the online adaptation\cite{shah2022offline}, the priority expert experience buffer\cite{lu1991distributed}\cite{sewak2019deep} could also speed up the training. In the ensuing part, we will present the results of our experiments, comparing the performance of our proposed framework with existing methods. Finally, we will conclude with a summary of our findings, a discussion on the potential implications of our research, and an overview of possible future directions in this exciting field of study.

The main contributions of proposed method are as follows:
\begin{itemize}
    \item We create a universal dataset for laser data navigation, which could be useful for the other people work.
    \item We propose a new pre-training offline DRL framework based on laser data.
    \item We design a training method for DRL, utilize the pre-training and expert experience to reduce training time.
    \item We obtain a general DRL policy model for avoidance navigation in most scenarios.
\end{itemize}
\section{Related Work}

Reinforcement Learning (RL) training is typically more time-consuming than supervised learning, largely due to the need for simulation environments to generate interaction data. In the early stages of RL, agents explore the environment via random actions to identify optimal paths, an approach that often results in superfluous actions and 'local convergence' - a state of being stuck in local optimum without improvement. These factors significantly extend the training duration and add complexity to the learning process.

In previous work, such as those conducted by Cheng et al. \cite{cheng2019end}, Yu Xiang al. \cite{zhang2022barrier} \cite{zhang2023adaptive} and Choi et al. \cite{choi2020reinforcement}, a constraint control barrier function or adaptive control\cite{liang2022adaptive}\cite{liang2023adaptive} was introduced as a novel solution to guide the exploration of the vehicle, ensuring it steers clear of obstacles and walls. This technique effectively shapes the action space explored by the agent, confining it to regions of interest. This strategic restriction not only enhances the safety of the vehicle's operations, but it also significantly reduces the training time by eliminating unnecessary exploration. The agent thus learns more efficiently, focusing its learning on actions that contribute to more optimal and safer navigation. 

Visual navigation, when integrated with Reinforcement Learning (RL), has emerged as a prevalent approach in the field of mobile robotics, as demonstrated by recent studies such as Wu et al. \cite{wu2021learn}, \cite{wu2018learn}, \cite{wu2019tdpp} and Yafei. \cite{hu2023off} in exploration in challenging environment. Typically, these methods utilize RGB images as the state for training an agent capable of collision-free navigation. This approach enhances a robot's ability to perceive its environment and enriches the semantic information available for learning, thereby enabling the agent to accumulate more nuanced experiences. 

Despite their advantages, this approach also has limitations. A single camera, for instance, cannot provide a full viewing angle, limiting the field of vision for the robot. There are also challenges associated with computational limitations of the robot and the 'sim-to-real' gap\cite{zhao2020sim}, a persistent issue where behaviours learned in simulation do not transfer perfectly to the real world. 

The work on \cite{kastner2023holistic} and \cite{cimurs2021goal} used global planning (some heuristic search method) and local planning (DRL) to achieve the goal oriented navigation. But our work focus on no collision local navigation and utilize the offline data collection and offline training to speed up the training.

Furthermore, the nature of RL, which requires real-time interaction with the simulation for training, does not fully leverage the advantages offered by parallel computing capabilities of modern GPUs. This contrasts with other areas of deep learning, which can fully benefit from these computational resources. Therefore, while visual navigation coupled with RL has shown promise, these considerations underscore the need for continued refinement and innovation in this domain.

\begin{figure*}[!htb]
\centerline{\includegraphics[width=\textwidth, height=0.3\textheight]{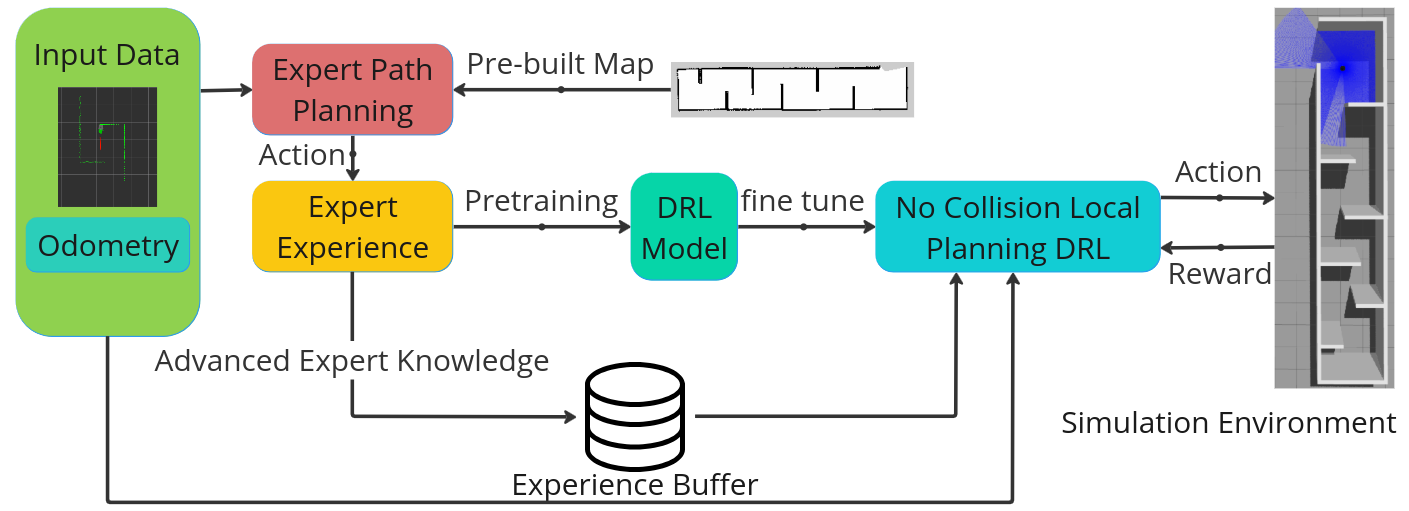}}
\caption{Framework of a system design. The input is exemplary for pretraining and fine tuning. The details of DRL algorithm and environment parameters are specified in section B and C.}
\label{structure}
\end{figure*}

\section{Background and Preliminary}
The goal of reinforcement learning is to discover a policy, denoted as $\pi$, that maximizes the expected cumulative discounted rewards. In the realm of Deep Reinforcement Learning (DRL), this policy $\pi_\theta$ is represented by a deep neural network, where $\theta$ refers to the tunable weights. This problem is formulated as a Markov Decision Process\cite{puterman1990markov} (MDP) with the following components: ${S, A, P, r}$. Here, $S$ is the state belonging to $\mathbb{R}^n$, $A$ is the action that belongs to $\mathbb{R}^m$, $P$ is the state transition function that maps from $S \times A \to S$, and $r$ is the reward function which maps from $S \times A \to \mathbb{R}$. In this context, we are tackling the problem in a continuous state and action space. 

Deep Deterministic Policy Gradient\cite{lillicrap2015continuous} (DDPG) is a classical actor-critic algorithm and combining the deterministic policy gradient approach to learn policy for continuous action spaces. 
Every time sample a mini-batch of $N$ transitions $(s_i, a_i, r_i, s_{i+1}, d_i)$ from buffer. Update the critic by minimizing the loss:
\begin{equation}
    L(\phi) = \frac{1}{N} \sum_i (Q_{\phi}(s_i, a_i) - (r_i + \gamma (1 - d_i) Q_{\phi'}(s_{i+1}, \pi_{\theta'}(s_{i+1}))))^2
\end{equation}
where $\gamma$ is the discount factor, $\pi_{\theta}$ is actor network, $Q_{\phi}$ is critic network, $\pi_{\theta'}$ is target actor network, $Q_{\phi'}$ is target critic network.
Update the actor using the sampled policy gradient:
\begin{equation}
    L(\theta)= \frac{1}{N} \sum_i Q_{\phi}(s_i, a)|_{a=\pi_{\theta}(s)}
\end{equation}
Update the target networks:
\begin{equation}
    \theta' \gets \tau\theta + (1 - \tau)\theta' 
\end{equation}
\begin{equation}
    \phi' \gets \tau\phi + (1 - \tau)\phi'
\end{equation}
where $\tau$ is the soft update rate.

Soft Actor-Critic\cite{haarnoja2018soft} (SAC) is a off-policy reinforcement learning algorithm that aims to maximize expected return while also maintaining exploration, which could also be used in continuous action space.
Similar to the DDPG main steps, SAC central feature is the entropy regularization, for update critic network first calculate the target Q-value, denoted as $y_i$:
\begin{equation}
    y_i = r_i + \gamma (1 - d_i) \left( \min_{j=1,2} Q_{\phi_j'}(s_{i+1}, a_{i+1}') - \alpha \log \pi_{\theta'}(a_{i+1}'|s_{i+1}) \right)
\end{equation}
where $\gamma$ is the discount factor, $Q_{\phi_j'}(s_{i+1}, a_{i+1}')$ is the Q-value of the next state and action, estimated by the target critic network, $\alpha$ is the temperature parameter, $\log \pi_{\theta'}(a_{i+1}'|s_{i+1})$ is the log-probability of the action under the current policy, $Q_{\phi_{1,2}}(s_i, a_i)$ is the Q-value of the current state and action, estimated by the critic network, then update the critic networks by minimizing the following loss:
\begin{equation}
    L(\phi_{1,2}) = \frac{1}{N} \sum_i (Q_{\phi_{1,2}}(s_i, a_i) - y_i)^2
\end{equation}
$N$ is the size of the mini-batch.
The actor is updated to maximize a trade-off between expected return and entropy, which encourages exploration:
\begin{equation}
    L(\theta) = \frac{1}{N} \sum_i \left( \min_{j=1,2} Q_{\phi_j}(s_i, \pi_\theta(s_i)) - \alpha \mathbb{E}_{\pi_\theta}[ \log \pi_\theta(a_i|s_i) ] \right)
\end{equation}

Twin Delayed Deep Deterministic Policy Gradient\cite{fujimoto2018addressing} (TD3) is an algorithm that builds upon the DDPG method by addressing function approximation errors and overestimation bias.It updates the policy (and target networks) less frequently than the Q-function and adds noise to the target action, to make it harder for the policy to exploit Q-function errors by smoothing out Q along changes in action.
Compute the target Q-value:
\begin{equation}
    y_i = r_i + \gamma (1 - d_i) \min_{j=1,2} Q_{\phi_j'}(s_{i+1}, \pi_{\theta'}(s_{i+1}) + \epsilon)
\end{equation}
Update the critics by minimizing the loss:
\begin{equation}
    L(\phi_{1,2}) = \frac{1}{N} \sum_i (Q_{\phi_{1,2}}(s_i, a_i) - y_i)^2
\end{equation}
Every after policy delay $d$ steps, update the actor by maximizing the Q-value:
\begin{equation}
    L(\theta) = \frac{1}{N} \sum_i Q_{\phi_1}(s_i, \pi_\theta(s_i))
\end{equation}
Softly update the target networks:
\begin{equation}
    \theta', \phi_{1,2}' \gets \tau\theta, \phi_{1,2} + (1 - \tau)\theta', \phi_{1,2}'
\end{equation}

\section{Methodology}

In this chapter, we outline the methodology behind our proposed framework, which is divided into two distinct stages. In the first stage, we gather data from experts, consolidate this data into a unified dataset, and use a Deep Reinforcement Learning (DRL) model to pretrain our model.

The second stage involves loading the pretrained model into our actor and critic networks to fine tune. We enrich the replay buffer with prioritized data, promoting an interactive learning process with the simulation. This dual-stage process not only enhances the training efficiency but also equips the agent with advanced expert knowledge(AEK), facilitating a more effective learning experience.

Figure \ref{structure} illustrates the system design of our approach. The collecting data and fine tuning process rely on ROS Gazebo simulation. The pre-built map could be obtained by the classical SLAM gmapping \cite{xuexi2019slam} lidar-based method. For the no collision local planning DRL training, the expert pretrained model will be loaded and the fine tune process will also sample some prioritized experience.

\subsection{Creating Pretraining Dataset}\label{AA}
During this process firstly we need build the map of environment via SLAM and develop an automatic tool to generate a goal in a limited area to collect the data what we need. The global path planning are A*\cite{stentz1994optimal} or DWA\cite{fox1997dynamic} algorithms. The data structure could be regarded as an MDP statement including state, action, reward, new state and reward. 

\begin{figure}[!htb]
\centerline{\includegraphics[width=0.45\textwidth]{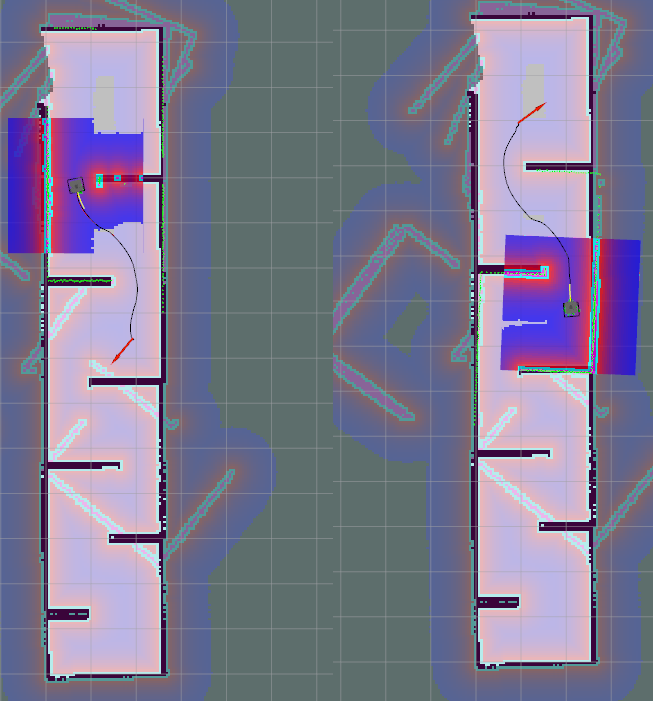}}
\caption{Global Path Planning by DWA.}
\label{Global Path Planning}
\end{figure}

\begin{itemize}
    \item Randomly generate a goal in the Gazebo environment, ensuring the goal's feasibility by avoiding placement on obstacles or walls.
    \item Employ the rostopic command to set the position and orientation through $/move base simple/goal$.
    \item Launch the AMCL (Adaptive Monte Carlo Localization\cite{xiaoyu2018adaptive}) node to perform localization and execute the global path planning algorithm in Figure \ref{Global Path Planning}.
    \item Collect data including linear and angular velocity as action, and discretized laser data coupled with odometry information as state. After a fixed interval of 0.5s, calculate the reward (see Equation \ref{reward} for details), verify if the vehicle has reached the goal or collided with a wall or obstacle, and gather the next state observation.
    \item Package the state, action, reward, new state, and done tuple into a tensor for efficient parallel pretraining.
\end{itemize}

\subsection{Pretraining}
Our mobile robots is defined on the Euclidean plane thus the laser data is 2D type. At each time step t, the robot has access to the surrounding observation state $s_t$ and generate the action velocity command $a_t$ that allows collision free navigation for the robot to the goal. This assumption makes our method abstract to train a policy function $f$ translated:
\begin{equation}
    a_t = f(s_t)
\end{equation}

\subsubsection{Definition of State and Action}
\begin{equation}
    \Vec{s_t} = \text{concat}(\Vec{l_t}, \Vec{g_t},  \Vec{a_{t-1}})
\end{equation}
The state, denoted as $s_t$ vector in 40 dimension, is composed of three distinct elements: the 36-point discretized laser data array $l_t$ which spans a full $360^{\circ}$ range, the goal position $g_t$ represented in polar coordinates, and the velocity $a_{t-1}$ from the previous time step.

The action space $a_t$ consists of linear velocity $v_t$ and angular velocity $\omega_t$
\subsubsection{Reward Function}
Given the requirement for the mobile robot to navigate towards the goal while maintaining a safe distance from obstacles, the reward function has been designed with careful consideration of both the clearance distance from obstacles and the incremental progress towards the goal. The reward function is formulated as follows:

\begin{equation}\label{reward}
r = r_{distance} + r_{collision} + r_{velocity} + r_{arrive}
\end{equation}

In this equation, the reward $r$ is a composite of several factors: $r_{distance}$ representing progress towards the goal, $r_{collision}$ accounting for obstacle avoidance, $r_{velocity}$ contributing the velocity factor, and $r_{arrive}$ signifying the successful arrival at the goal. The term $r_{distance}$ is computed as follows:

\begin{equation}
r_{distance} = w_{distance}*(d_{t-1} - d_t)
\end{equation}

where $d_{t-1}$ signifies the distance to the goal from the previous time step, while $d_t$ represents the current distance to the goal $w_{distance}$ is hyper-parameter. The reward $r_{distance}$ thus captures the change in distance towards the goal between consecutive time steps. The reward collision $r_{collision}$ can be calculated by:

\begin{equation}
    r_{collision} = 
    \begin{cases} 
      0 & \text{if } \min(l_t) > 2d_{threshold} \\
      r_c & \text{if } d_{threshold} < \min(l_t) \leq 2d_{threshold} \\
      2 * r_c & \text{if } \min(l_t) \leq d_{threshold} 
    \end{cases}
\end{equation}

Where, $l_t$ represents laser data vector, $r_c$ is the punishment of collision and the $d_{threshold}$ is the threshold value of distance between robot and obstacle. The term $r_{velocity}$ is computed as follows:

\begin{align}
    r_{velocity} &= r_{linear} + r_{angular} \\
    r_{linear} &= 
    \begin{cases} 
      0 & \text{if } v_t \geq 0.1 \\
      -4 & \text{if } v_t < 0.1
    \end{cases} \\
    r_{angular} &= 
    \begin{cases} 
      0 & \text{if } -0.5 \leq \omega_t \leq 0.5 \\
      -1 & \text{if } \omega_t > 0.5 \text{ or } \omega_t < -0.5
    \end{cases}
\end{align} 

In these equations, $v_t$ and $\omega_t$ represent the linear and angular components of the action vector respectively. The rewards $r_{linear}$ and $r_{angular}$ are calculated based on these components. They are then combined to compute the overall velocity reward $r_{velocity}$. This design could keep the vehicle go ahead as far as possible rather than turning around. The term $r_{arrive}$ is computed as follows:

\begin{equation}
    r_{arrive} = 
    \begin{cases} 
      R_{success} & \text{if done and arrived} \\
      R_{fail} & \text{if done and crushed} \\
      0 & \text{otherwise}
    \end{cases}
\end{equation}

In this equation, "done" is a boolean variable that becomes True when an episode ends (either by reaching the goal or due to a collision).

\subsubsection{Network architecture}
\begin{figure*}[!htb]
\centerline{\includegraphics[width=\textwidth, height=0.3\textheight]{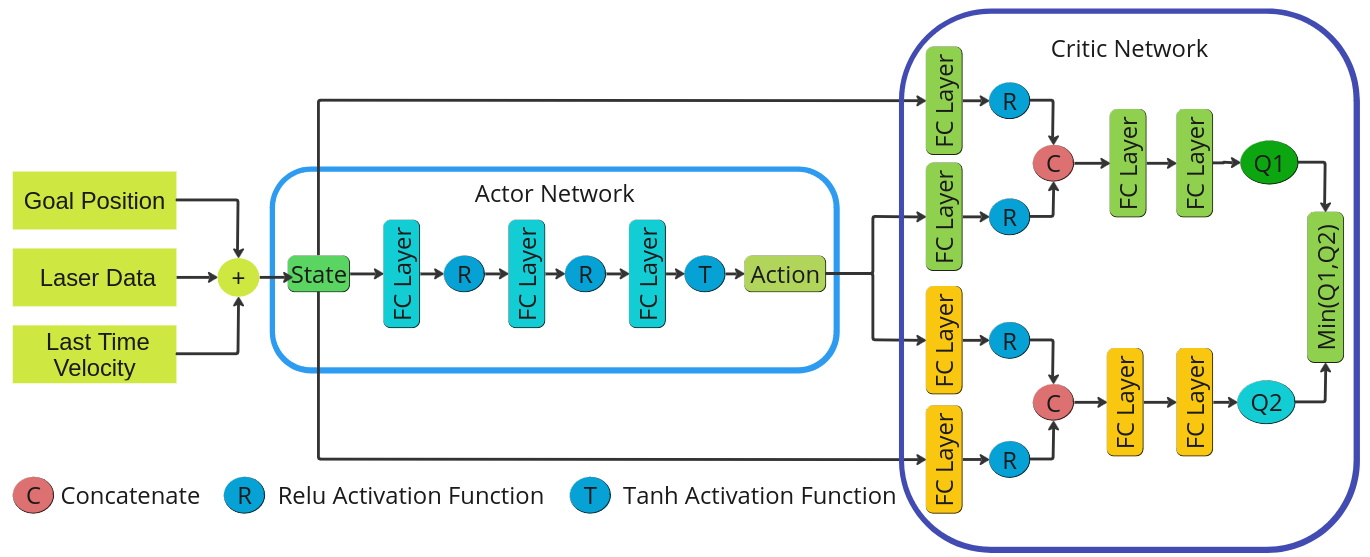}}
\caption{The architecture of DRL network, Actor and Critic Network.}
\label{Network}
\end{figure*}

In this study, we leverage the Deep Deterministic Policy Gradient (DDPG), Soft Actor-Critic (SAC), and Twin Delayed Deep Deterministic Policy Gradient (TD3) algorithms as the foundational architecture of our proposed model. As depicted in Fig. \ref{Network}, the actor network ingests a 40-dimensional input vector, which fuses a 36-dimensional laser data, 2-dimensional goal coordinates in polar form, and the action from the previous time step. This data is processed through three fully connected layers, each containing 256 nodes, ultimately generating linear and angular velocities via a Tangh function.

The output action is amalgamated with the state to serve as the input for the critic network, which mirrors the actor network in terms of hidden nodes and features two independent networks. However, the critic network introduces a unique aspect where the action and state are individually channeled into a fully connected layer, each producing a vector with half the number of hidden nodes. These vectors are then concatenated into a single vector with the same width as the hidden layer. Post-processing through two additional fully connected layers enables the two independent critic networks to generate $Q_1$ and $Q_2$ respectively.

\subsubsection{Overall Algorithm}
The overall algorithm is then presented. Algorithm \ref{algorithm offline} describes offline pre-training with expert experience dataset collected above.

\begin{algorithm}
\caption{Offline Pre-Training}
\label{algorithm offline}
\begin{algorithmic}[1]
\REQUIRE A collected expert experience dataset.
\REQUIRE Training expert trajectories dataset: $D_{tr} = \{s_i,a_i,r_i,s_{i+1},dw_i\}$, representing state, action, reward, next state, done.
\ENSURE Learned policy function to map state to action and critic functions: $\hat{\pi}(s, \theta), \hat{Q}_1(s, a, \phi_1), \hat{Q}_2(s, a, \phi_2), \forall s \in S$
\STATE Initialize actor and critic network weights $\theta$, $\phi_1, \phi_2$ and set policy delay $\delta$.
\FOR{each training epoch}
    \STATE Randomly sample a batch of transitions, $B = \{(s_i,a_i,r_i,s_{i+1},d_i)\}$
    \STATE Compute target Q-value
    \STATE Update critics by minimizing the loss
    \IF{$j$ mod $\delta$ $= 0$}
        \STATE update the actor by maximizing the Q-value
    \ENDIF
    \STATE Update target network
\ENDFOR
\end{algorithmic}
\end{algorithm}
Remark: The motivation of offline pre-training is to
\begin{itemize}
    \item Reduce the exploration of online adaption.
    \item Introduce sophisticated expert knowledge that is challenging for the DRL agent to discover on its own.
\end{itemize}

\subsection{Online Training for Local Planning}
\begin{figure}[!htb]
\centerline{\includegraphics[width=0.3\textwidth]{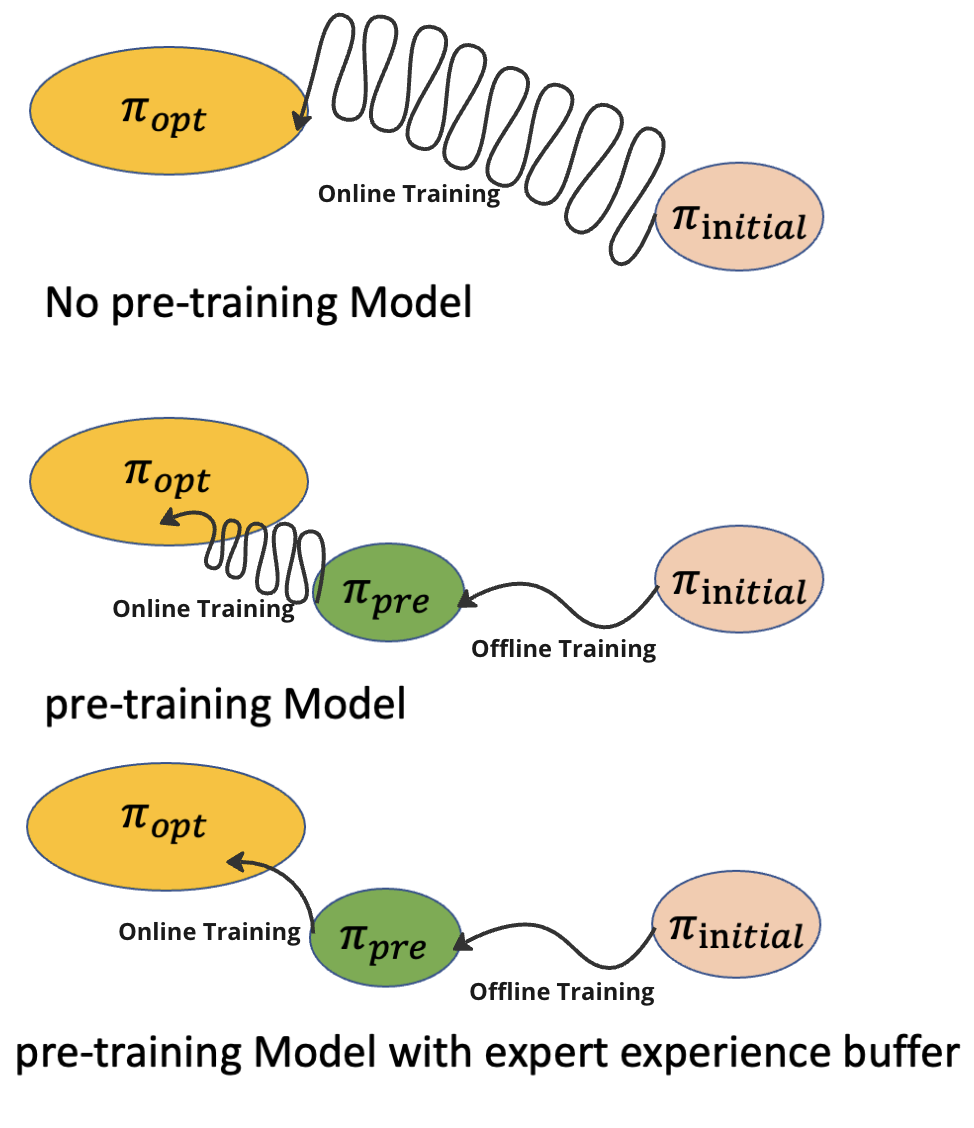}}
\caption{Pre-Training And Expert Experience Buffer}
\label{Pre-training}
\end{figure}

Instead of initializing the actor and critic network randomly, the online training for adaption with interaction of gazebo simulation could load the pre-training model to reduce the training time. At the same time, the expert experience could also be added in the replay buffer as a kind of supervised learning approach. Our target is training a no collision policy $\pi_{opt}$, the pre-training policy is $\pi_{pre}$, showed in Fig. \ref{Pre-training}
\begin{equation}
    \pi_{online} = \pi_{target} - \pi_{pre}
\end{equation}
When the more $\pi_{pre}$ close to $\pi_{opt}$, the training time could be highly reduced.

Algorithm \ref{algorithm online} describes online DRL learning and testing. Note that we asume $\theta$ and $\theta'$ to denote the network weights for pre-training and testing, respectively, dataset $D_{tr}$ and $D_{te}$, etc. 

\begin{algorithm}
\caption{Online Adaptation}
\label{algorithm online}
\begin{algorithmic}[1]
\REQUIRE State representation laser data and action velocity.
\REQUIRE Pre-training network weights: $\theta$, $\phi_1$, $\phi_2$
\REQUIRE Fine-tuning Trajectory data $D_{te} = \{s_i,a_i,r_i,s_{i+1},dw_i\}$ in online simulation.
\ENSURE Fine-tuned policy function to map state to action and critic functions: $\hat{\pi}(s, \theta'), \hat{Q}_1(s, a, \phi'_1), \hat{Q}_2(s, a, \phi'_2), \forall s \in S$
\STATE Initialize online network weights $\theta' = \theta$, $\phi'_1 = \phi_1$, $\phi'_2 = \phi_2$ and set policy delay $\delta$.
\FOR{each training epoch}
    \STATE Receive state observation $s_t$ from simulation
    \STATE Calculate the action $a_t$ following $\pi(a_t|s_t)$
    \STATE Implement action $a_t$ in simulation and wait for the fixed time step.
    \STATE Compute the reward $r_t$, get the next observation $s_{t+1}$ and judge whether reached or crushed $d_t$.
    \STATE Store $\{s_t,a_t,r_t,s_{+1},d_t\}$ in buffer $D_{te}$
    \IF{$j$ mod $\delta$ $= 0$}
        \STATE Update the critic and actor network weight $\theta, \phi_1, \phi_2$
    \ENDIF
    \IF{$dw_t$ = True}
        \STATE Rest the simulation
    \ENDIF
\ENDFOR
\end{algorithmic}
\end{algorithm}

\section{Experiments}
\subsection{Pre-Training and experiment setup}
In our simulation environment, we employed the TurtleBot3 mobile robot, as depicted in Fig. \ref{structure}. The process of dataset collection was executed within the Gazebo environment. This approach significantly eases the transfer of acquired expertise from the simulation to real-world robot experiments, as the simulated and real models utilize identical ROS interfaces and the Gazebo world also accurately emulates the robot's dynamics.

Our robot's pre-training dataset was compiled within a confined corridor environment. During this collection phase, the goal target was arbitrarily positioned within the corridor landscape to enhance the model's generalization capabilities. The surrounding laser information is represented by blue lines. The robot is subscribed to laser data, providing a comprehensive $360^{\circ}$ field of view with a range extending from 0.1m to 6.5m. For the purposes of this study, we have assumed that the robot will solely move forward.

The pre-training was performed on a personal computer, equipped with an NVIDIA RTX 1660Ti. The model was trained using the Adam optimizer. \cite{zhang2018improved}.

\subsection{Pre-training evaluation}
\begin{figure}[!htb]
\centerline{\includegraphics[width=0.5\textwidth]{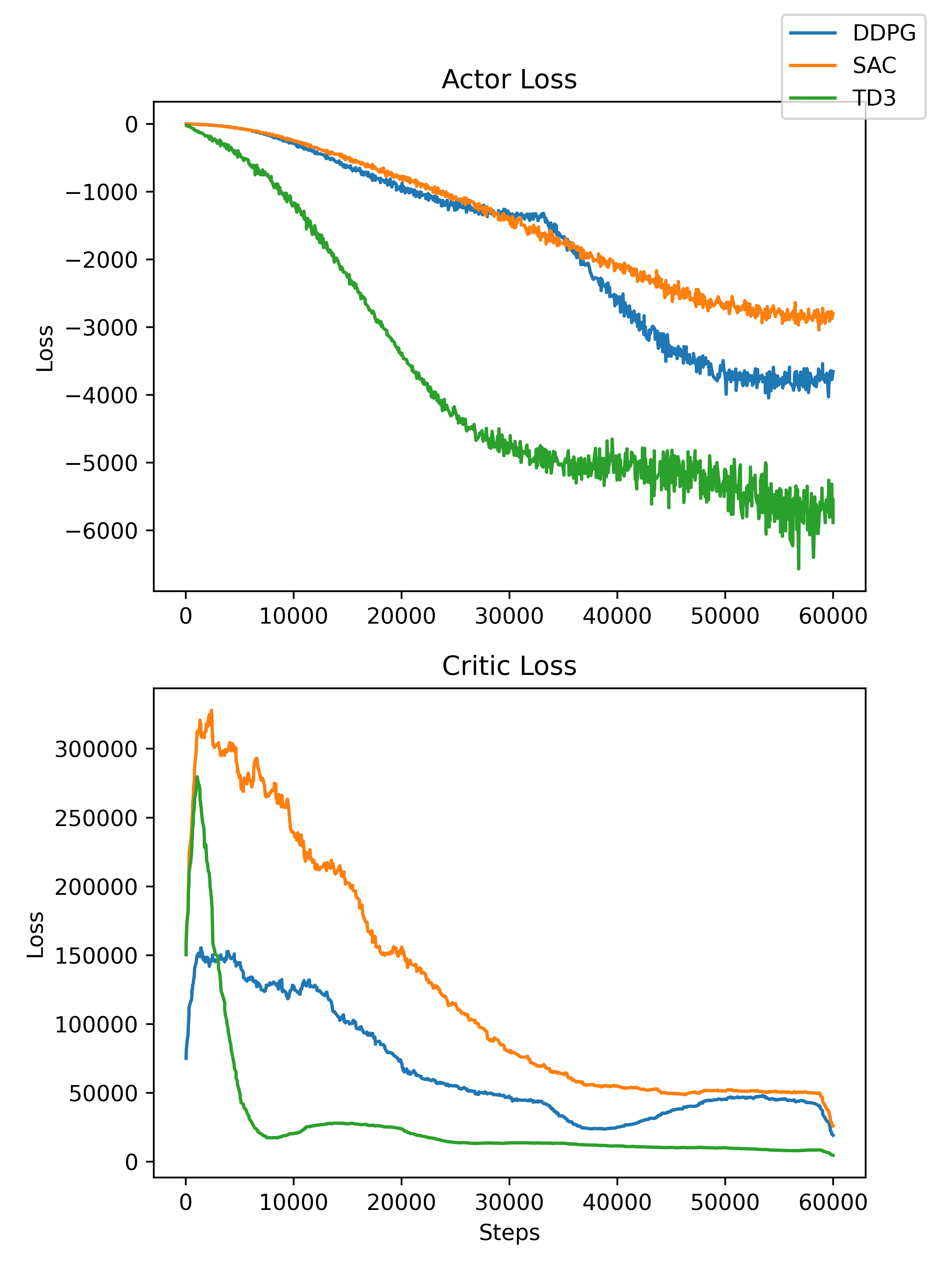}}
\caption{Pre-Training for Actor and Critic Network Loss}
\label{Offline_loss}
\end{figure}

We tried three standard DRL algorithms (DDPG, SAC and TD3) were implemented for comparision, shown in Fig. \ref{Offline_loss}. At the same time the details about training details is in Table. \ref{tab1}. In a comparative study of the DDPG, SAC, and TD3 pre-training, significant differences were observed in their performances. The TD3 algorithm notably outperformed the other two, boasting the lowest actor loss among all three. More importantly, TD3 demonstrated a swift convergence, attaining its optimal performance in approximately 10,000 steps. This rapid convergence of TD3 is a significant advantage, especially when compared to the DDPG and SAC algorithms, which required approximately 40,000 steps to converge. This efficiency of TD3 can be pivotal in applications where time is a critical factor. Thus, in terms of both actor loss and rate of convergence, TD3 proves to be the superior algorithm among the ones compared.

\begin{table}[htbp]
\caption{Pre-Training comparison}
\begin{flushleft}
\begin{tabular}{|c|c|c|}
\hline
\textbf{Run} & \textbf{Min, Max, End Value} & \textbf{Steps} \\
\hline
DDPG & -3,811.389, -0.063, -3,733.363 & 59,999 \\
\hline
SAC & -2,865.133, -0.242, -2,839.005 & 59,999 \\
\hline
TD3 & -5,819.918, -25.101, -5,663.639 & 59,999 \\
\hline
\end{tabular}
\end{flushleft}

\begin{flushleft}
\begin{tabular}{|c|c|c|c|}
\hline
\textbf{Run} & \textbf{Learning Rate} & \textbf{Batch Size} & \textbf{End Time} \\
\hline
DDPG & 3e-6 & 256 & 5.663 min \\
\hline
SAC & 3e-6 & 256 & 7.9 min \\
\hline
TD3 & 3e-6 & 256 & 4.615 min \\
\hline
\end{tabular}
\label{tab1}
\end{flushleft}
\end{table}

\subsection{Online adaption}
\begin{figure}[!htb]
\centerline{\includegraphics[width=0.5\textwidth]{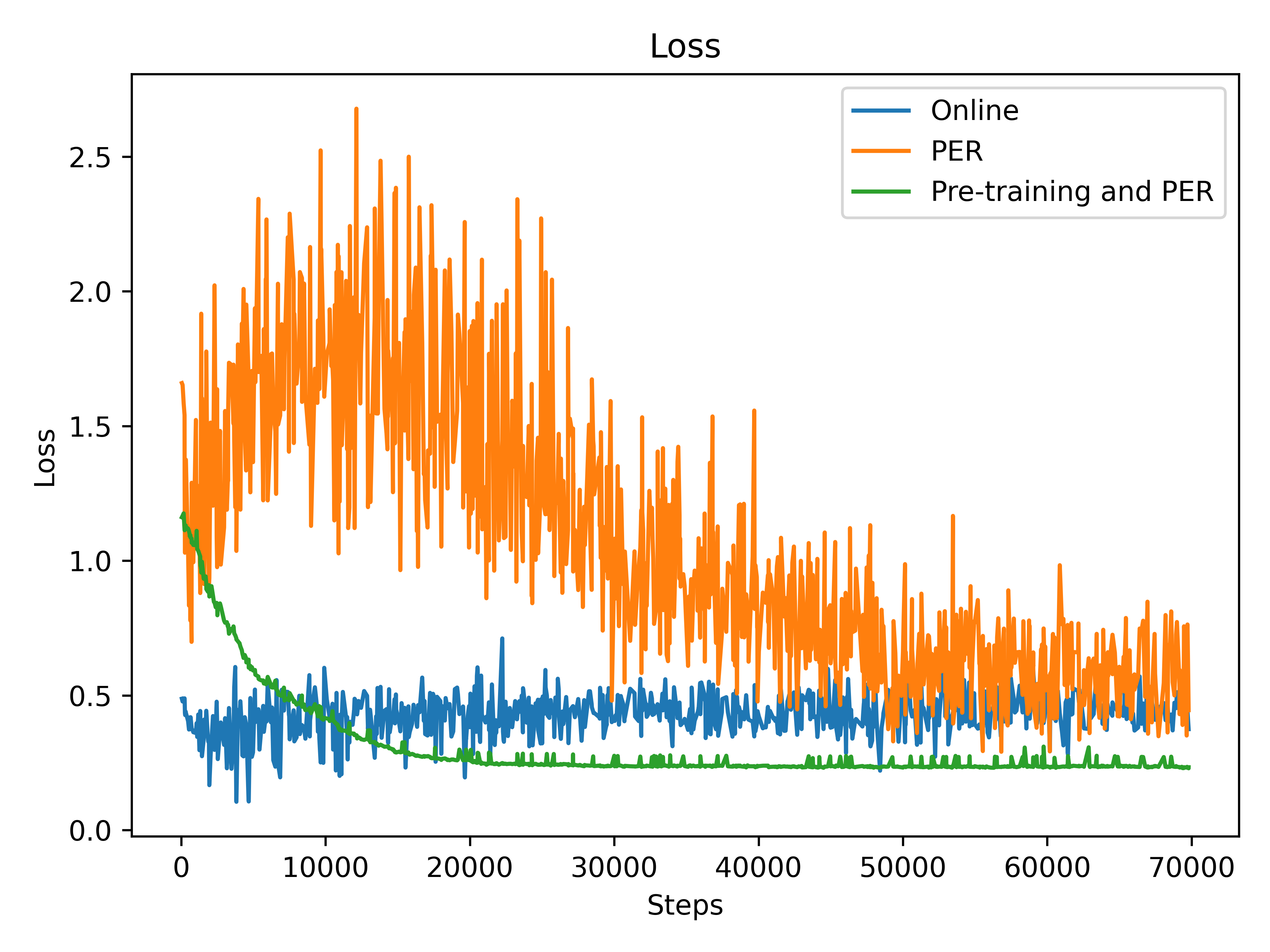}}
\caption{The online training loss for Comparison on pre-training and Prioritized Expert Experience Replay}
\label{online_loss}
\end{figure}

After conducting trials with three distinct online adaptation methods - Online Learning, Prioritized Expert Experience, and Pre-training - the training loss curve illustrated in Fig. \ref{online_loss} indicates that the Pre-training approach converges swiftly. In comparison to models without pre-training, the latter exhibit slower convergence and lack stability. If we resort to direct online training, the loss remains constant and does not diminish, suggesting ineffective training that leads to immediate local optimum trapping. This disparity underscores the efficacy and superiority of the pre-training method in promoting faster convergence and overall model stability.

\begin{figure}[!htb]
\centerline{\includegraphics[width=0.5\textwidth]{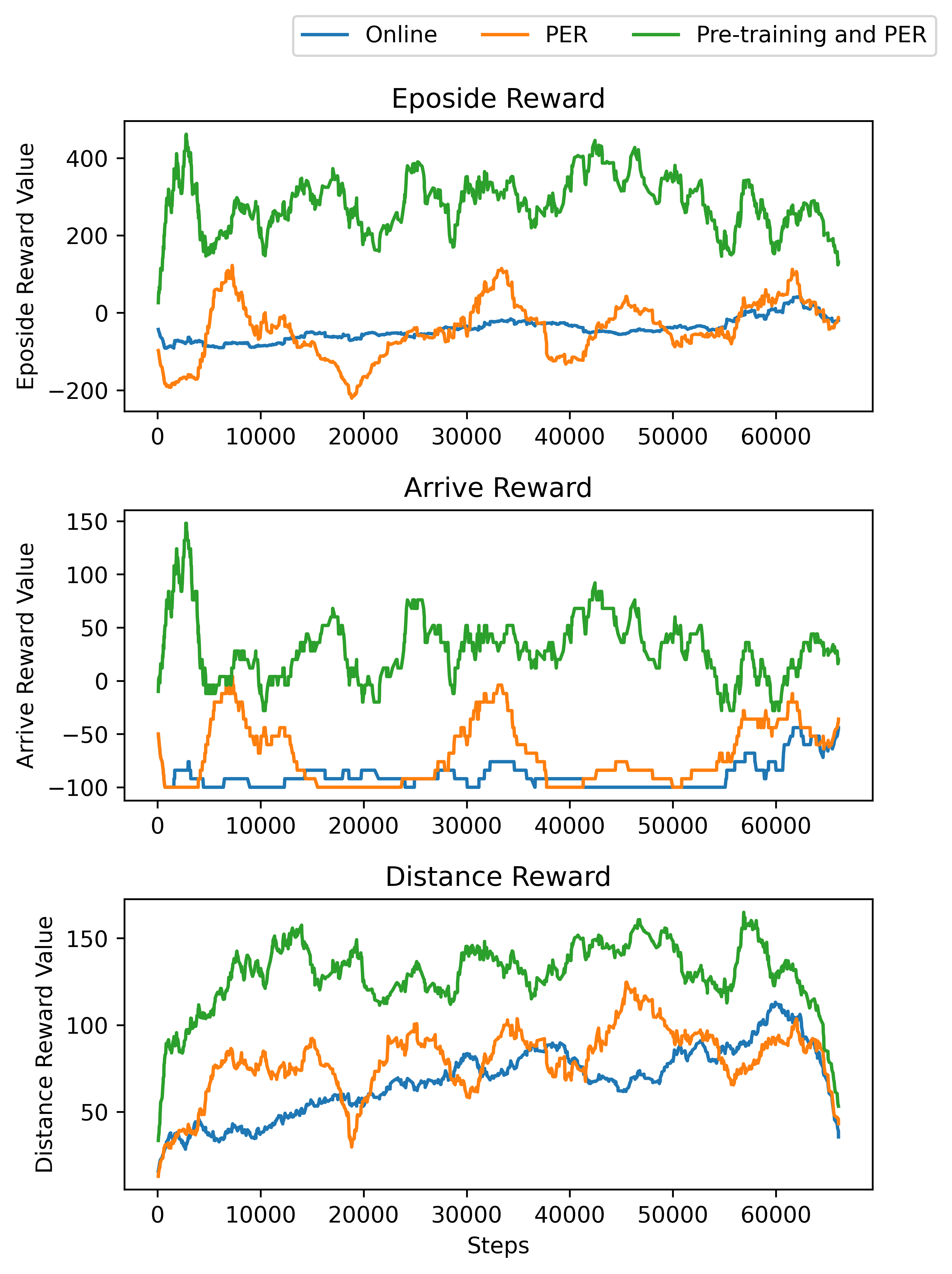}}
\caption{The online training Rewards for Comparison on pre-training and Prioritized Expert Experience Replay}
\label{online_reward}
\end{figure}

Figure \ref{online_reward} compares the episode reward, arrival reward, and distance reward. These rewards respectively represent the total reward in one epoch, whether the goal is reached, and the changes in distance toward the goal at consecutive time steps. Evidently, the pre-training approach augmented with prioritized expert experience outperforms the other two methods, securing substantially higher rewards. The primary contrast in this comparison is that the other two methods fail to show a significant increase in goal arrivals, attributable to the absence of guided initialization. The evidence thus strongly suggests the superiority of the pre-training method combined with prioritized expert experience in enhancing the performance metrics.

\section{Conclusion and Future Work}

In this paper, we proposed a methodology for offline pre-training based on DRL for collision avoidance. Our approach capitalizes on the advanced expert experience pre-training to mitigate inefficient random explorations typically encountered during the early stages of online interactive training. In a simulated corridor environment, our method demonstrated efficiency, requiring only 20\% of the training steps compared to the Prioritized Expert Experience Replay (PER) method. In contrast, standard DRL methods were seen to fall into local optima and struggled to enhance rewards. Our methodology only took less than 4 hours of training time to obtain superior rewards, while both PER and standard methods needed more than two days to achieve comparable results.

It's important to note that our training objective was primarily to avoid collisions in local planning, resulting in less optimal global planning when relying solely on DRL. This can lead to certain invalid actions and increased costs for extended global navigation. The proposed method was also rigorously tested in various unknown simulations such as indoor houses, offices, and mazes, without any need for fine-tuning. The results suggest that our method exhibits robust and superior avoidance capabilities.

Looking ahead, we plan to explore the utilization of RGB images combined with depth perception and 3D point cloud data for training across a wider variety of tasks. Moreover, we are keen to integrate the transformer model \cite{vaswani2017attention} to enhance navigation based on attention mechanisms. Lastly, we aim to apply large-scale unsupervised model training for fine-tuning various downstream tasks.



\bibliography{ref}

\vspace{12pt}

\end{document}